# Deep learning models are not robust against noise in clinical text

Milad Moradi[1]

Institute for Artificial Intelligence and Decision Support, Center for Medical Statistics, Informatics, and Intelligent Systems, Medical University of Vienna, Vienna, Austria

`milad.moradivastegani@meduniwien.ac.at`

Kathrin Blagec

Institute for Artificial Intelligence and Decision Support, Center for Medical Statistics, Informatics, and Intelligent Systems, Medical University of Vienna, Vienna, Austria

`kathrin.blagec@meduniwien.ac.at`

Matthias Samwald

Institute for Artificial Intelligence and Decision Support, Center for Medical Statistics, Informatics, and Intelligent Systems, Medical University of Vienna, Vienna, Austria

`matthias.samwald@meduniwien.ac.at`

---

[1] Corresponding author. **Postal address:** Institute for Artificial Intelligence and Decision Support, Währinger Straße 25a, 1090 Vienna, Austria. **Telephone number:** 0043-1-40160-36313

# Abstract

Artificial Intelligence (AI) systems are attracting increasing interest in the medical domain due to their ability to learn complicated tasks that require human intelligence and expert knowledge. AI systems that utilize high-performance Natural Language Processing (NLP) models have achieved state-of-the-art results on a wide variety of clinical text processing benchmarks. They have even outperformed human accuracy on some tasks. However, performance evaluation of such AI systems have been limited to accuracy measures on curated and clean benchmark datasets that may not properly reflect how robustly these systems can operate in real-world situations. In order to address this challenge, we introduce and implement a wide variety of perturbation methods that simulate different types of noise and variability in clinical text data. While noisy samples produced by these perturbation methods can often be understood by humans, they may cause AI systems to make erroneous decisions. Conducting extensive experiments on several clinical text processing tasks, we evaluated the robustness of high-performance NLP models against various types of character-level and word-level noise. The results revealed that the NLP models' performance degrades when the input contains small amounts of noise. This study is a significant step towards exposing vulnerabilities of AI models utilized in clinical text processing systems. The proposed perturbation methods can be used in performance evaluation tests to assess how robustly clinical NLP models can operate on noisy data, in real-world settings.

# 1. Introduction

Recent years have witnessed major breakthroughs in developing Artificial Intelligence (AI) systems mastering complicated tasks that require human intelligence. Deep Learning (DL) is a subfield of AI that allows intelligent systems to learn complicated data relationships, encode the learned knowledge into numerical representations, and use the representation model to build accurate predictive models [1]. DL methods have received much attention for developing computational models in bioinformatics and clinical informatics research [2-6]. More specifically, DL methods have been widely applied to clinical text data such as medical and health records, clinical notes, diagnostic notes, pathology reports, discharge summaries, nursing documentation, etc. for a plethora of tasks including medical concept extraction, phenotyping, clinical relation extraction, cancer-case identification, clinical and translational research, clinical decision support, and drug-related studies [7-9]. Powerful DL models have shown state-of-the-art performance in clinical Natural Language Processing (NLP), even surpassing human accuracy on some clinical text processing benchmarks [10-13].

However, recent studies have raised concerns about the gap between the performance on benchmark datasets and the ability of NLP systems to effectively operate in real-world situations [14, 15]. Clinical NLP models are usually evaluated using clean and human-curated datasets not properly representing diverse, heterogeneous, and noisy inputs an AI system may face in real-world use-cases. Clinical texts may have no clear structure and use various communication styles [16]. Moreover, noisy inputs, misspelled words, grammatical errors, eliding words that should be inferred from context, duplication, excessive use of ad-hoc abbreviations, acronyms, and jargon are among common challenges an automatic clinical text processing system should properly deal with [16-19]. In addition to evaluation on benchmark datasets, AI systems developed for clinical text processing should be also evaluated on noisy inputs that resemble characteristics of clinical texts in real-world settings. This helps to have a more balanced estimation of the performance, robustness, and reliability of AI systems that are meant to be used in a wide variety of clinical settings.

In this paper, we investigate how robust clinical AI systems are against noisy inputs in text processing tasks. We designed and implemented various perturbation methods that simulate different types of noise in clinical text data. Conducting extensive experiments on four different clinical NLP tasks, we evaluated the ability of state-of-the-art NLP models in handling noisy inputs. The perturbation methods applied minor alterations to text to resemble common types of noise in clinical text such as misspelled words, grammatical errors, missing words, duplication, etc. We also examined how well the NLP models can handle abbreviations and acronyms, synonym words, upper case/lower case words, and other sorts of variability in text. Our experimental results revealed that the NLP systems are sensitive to noisy data that can be easily handled by human. The main contributions of this paper are as follows:

1) We introduce and implement a wide variety of character-level and word-level perturbation methods

that simulate different types of noise a clinical text processing system may encounter in real-world use cases.

2) We evaluate the robustness of state-of-the-art NLP models against noisy inputs on various clinical text processing tasks.

3) We investigate the usefulness of the automatically generated perturbations through an extensive user study.

To the best of our knowledge, this work is the first to evaluate and compare the robustness and reliability of multiple state-of-the-art, deep-learning clinical NLP models across several clinical text processing tasks. Compared to the CheckList testing tool [14], our perturbation methods do not produce synthetic samples and use the test set's input space for generating new test samples. It also does not need persistent user intervention. Furthermore, our perturbation methods aim at testing both robustness and linguistic capabilities, while CheckList focuses on the latter. Most of perturbation-based evaluation methods rely on adversarial attacks [20, 21] that require access to NLP system's internal weights and model structures in order to generate effective adversarial examples. However, having access to a system's internals is not always feasible in every test scenario. On the other hand, our perturbation methods treat NLP systems as black-boxes and can therefore be applied to any system regardless of its underlying AI model, parameters, or the target NLP task. This work can be a significant step forward in evaluating medical intelligent systems in realistic settings. This study brings more insights into vulnerabilities of AI systems in real-world situations, and suggests that performance evaluation on benchmark datasets should be complemented by tests that disclose regions of input space in which AI systems fail to operate accurately.

## 2. Clinical NLP models

High-performance neural NLP models leverage transfer learning [22] from massive amounts of unlabeled text to master linguistic features and transfer the knowledge to downstream applications. They are first pretrained on extremely large text corpora in an unsupervised manner to encode lexical, syntactic, and semantic properties of the language. They are then fine-tuned on downstream supervised tasks to learn task-specific language capabilities. BERT, XLNet, and ELMo are among powerful deep neural language models that take advantage of unsupervised pretraining and transfer learning; they have obtained state-of-the-art results on various NLP tasks. Domain-specific versions of these models have been developed as well, aiming at building AI systems for biomedical and clinical text processing.

We utilized three deep neural language models, i.e. ClinicalBERT, ClinicalXLNet, and ClinicalELMo, in our experiments to evaluate the robustness of high-performance AI models against input noise in clinical text processing. **ClinicalBERT** [12] was initialized from BioBERT (a variation of BERT pretrained on PubMed abstracts and PMC full-texts) and further pretrained on two million clinical notes from the MIMIC-III corpus. **ClinicalXLNet** [13] was initialized from the open-domain XLNet model, then was further pretrained on clinical notes from the MIMIC-III corpus. **ClinicalELMo** used the ELMo contextualized language modelling [23] as the base model and was pretrained on the MIMIC-III clinical texts.

## 3. Clinical text processing tasks

We evaluated the AI models on four different clinical text processing tasks. Table 1 summarizes some statistics of the datasets. We briefly introduce the datasets in the following.

**EBM PICO** [24] is a Named Entity Recognition (NER) dataset that contains almost 5000 clinical trial abstracts annotated for identifying text spans that describe populations, interventions, comparators, and outcomes.

**i2b2-2010** [25] is a Relation Extraction (RE) dataset that consists of more than 400 patient reports annotated for identifying eight types of relations between medical problem and treatment entities in the text.

**MedNLI** [26] is a Textual Inference (TI) dataset that contains a collection of sentence pairs from the MIMIC-III clinical notes dataset. Every pair is classified into one of the classes 'entailment', 'contradiction', or 'neutral' that specify whether the first sentence can be inferred from the second one or not.

**ClinicalSTS-2019** [27] consists of more than 2,000 pairs of sentences from the Mayo Clinics clinical text database, with scores between 0 and 5 assigned for the Semantic Similarity (SS) estimation task. A similarity score of 5 refers to semantic equivalence between a pair of sentences, while 0 signifies complete independence.

**Table 1.** Main statistics of the datasets used to evaluate the AI text processing models.

| Dataset | Task | Evaluation metric | Train | Dev | Test |
| --- | --- | --- | --- | --- | --- |
| EBM PICO | NER | F1 entity-level | 4300 | 500 | 200 |
| i2b2-2010 | RE | Micro F1 | 3000 | 121 | 6293 |
| MedNLI | TI | Accuracy | 11232 | 1395 | 1422 |
| ClinicalSTS-2019 | SS | Pearson | 1442 | 200 | 412 |

The evaluation metrics used to assess the performance of the language models on the different tasks are given in Table 1. *Accuracy* is computed as the proportion of correctly predicted samples to all samples in the dataset. *F1 score* is defined as the harmonic mean of Precision and Recall that are computed for every class in a dataset; Precision quantifies how precise a system can classify samples that belong to a class, while Recall quantifies how well the system can identify all samples that belong to the class. *Micro F1* computes an overall F1 score over all classes in a dataset such that every class contributes to the overall score in proportion to its size. The entity-level version of F1 score is defined as the average of F1 scores over all unique entities in the NER task. The *Pearson coefficient* quantifies the correlation between ground-truth scores assigned to samples and scores estimated by an AI system in a semantic similarity estimation task.

## 4. Noisy inputs

We designed and implemented seven character-level and nine word-level perturbation methods that simulate real-world situations in which clinical texts contain different types of noise. Table 2 presents examples of every character-level and word-level perturbation. The source code and other resources are available at https://github.com/mmoradi-iut/ClinicalTextPerturbation.

### 4.1. Character-level noise

**Deletion**. A word with more than two characters is selected in a random manner, and a character is removed from the word. The first and last characters of a word are never removed.

**Insertion**. A word is selected, and a random character is inserted in a random position, except in the first and last positions.

**Letter case changing**. The letter case of either the first character or all characters of a word is toggled, i.e. lower-case characters are replaced with their upper case form, and vice versa.

**Common misspelled words**. Words in the input are replaced with their misspellings from the Wikipedia corpus of common misspelled words.

**Repetition**. A character of the word is randomly selected and a copy of it is inserted right after the character.

**Replacement**. A character of the word is randomly selected and is replaced by an adjacent character on the keyboard.

**Swapping**. A character of the word is selected and swapped with its adjacent left or right character in the word.

### 4.2. Word-level noise

**Replacement with abbreviation**. Phrases in the input are replaced with their abbreviations (or acronyms) from a list of common medical and clinical abbreviations and acronyms [28-31].

**Abbreviation expansion**. Abbreviations and acronyms in the input are expanded to the phrases they refer to, using the list of common medical and clinical abbreviations and acronyms.

**Deletion**. A word is randomly selected and removed from the input.

**Negation**. Sentences of the input are negated by converting positive verbs to negative; or negation is removed from the sentences by converting negative verbs to positive. The goal is to investigate how well the language models can handle negation and properly reflects it in their decisions.

**Word order**. *M* consecutive words in the input are selected, and their order of appearance is changed. *M* is specified in a random manner from the range [2, |input|−1], where |input| denotes the number of words in the input. The goal is to investigate whether the language models are sensitive to the order in which the input words appear or they only care about the presence of important words regardless of their position in the input.

**Repetition**. A word is randomly selected, and a copy of it is inserted right after the word.

**Replacement with synonym**. Words in the input are replaced with their synonyms from a list of frequent medical and clinical words extracted by the SketchEngine language tool [32]. This list was created by collecting most frequent words in all the datasets used in the experiments. SketchEngine was then used to approximate similarities between the words and find synonyms through distributional semantic modelling.

**Singular/plural verb**. After tokenizing the input and assigning part of speech tags to the words, the subject is identified and the corresponding verb is converted to its plural form if the subject is singular, or it is converted to its singular form if the subject is plural. This perturbation simulates the situation in which a person who writes the text mistakenly uses a plural verb for a singular subject, or vice versa.

**Verb tense**. After tokenization and part of speech tagging, the subject is identified and the corresponding verb is converted to its past tense form if it appears in present tense, or it is converted to its present tense form if it appears in past tense.

As can be seen in Table 2, most of the perturbation methods do not change the original text's meaning, and the resulting noisy text is still understandable and meaningful. On the other hand, some word-level perturbations such as **Deletion**, **Negation**, **Word order**, and **Replacement with synonym** can change the input text's meaning such that the noisy text may be no longer meaningful, or it may become inconsistent with the ground-truth label. In this case, the label should be properly changed before testing the NLP system, to preserve consistency, or the sample should be excluded from experiments. In order to address this issue, we monitored every noisy sample resulting from the four above-mentioned perturbation methods and changed the ground-truth label if it was inconsistent with the perturbed text. We excluded those noisy samples that were no longer meaningful or understandable. Since monitoring, correcting, and filtering all the samples was highly time-consuming, we did this process until 200 samples were collected for each perturbation method on every dataset. For the rest of the perturbation methods, i.e. all the character-level ones and those word-level perturbations that are not likely to change the text's meaning with respect to the NLP tasks, we did not filter the noisy samples and used all of them in the experiments.

Table 2. Examples of character-level and word-level noise.

| Perturbation | Original text | Noisy text |
|---|---|---|
| *Character-level perturbation* | | |
| **Deletion** | Patient was agreeable to speaking with social work. | Patient was agreeable to speakng with social work. |
| **Insertion** | These were reviewed and reconciled with the patient, patient's spouse. | These were reviewed and reconciled with the pataient, patient's spouse. |
| **LCC** | Flonase 2 sprays to each nostril once daily as needed for nasal congestion. | Flonase 2 sprays to each nostril once daily as needed for NASAL congestion. |
| **CMW** | The patient denies a history of pacemaker or defibrillator. | The patient denies a history of pacemkaer or defibrillator. |
| **Repetition** | The patient has severe abdominal pain. | The patient has severe abdomminal pain. |
| **Replacement** | His preoperative work-up revealed no evidence of metastasis from his hepatoma. | His preoperative work-up revealed no evidence of metastasos from his hepatoma. |
| **Swapping** | 49 years-old male found to have a heart murmur a few months ago. | 49 years-old male fuond to have a heart murmur a few months ago. |
| *Word-level perturbation* | | |
| **RWA** | He denies any shortness of breath or difficulty breathing. | He denies any SOB or difficulty breathing. |
| **AE** | The patient has symptoms of a GI condition. | The patient has symptoms of a gastrointestinal condition. |
| **Deletion** | That day, he was found to be acutely short of breath with a respiratory rate of 40. | That day, he was found to be short of breath with a respiratory rate of 40. |
| **Negation** | She subsequently developed hypotension with SBP in the 70s. | She subsequently did not develop hypotension with SBP in the 70s. |
| **Word order** | This therapist fit the patient with the orthosis listed in the Treatment Plan. | This therapist fit the patient with the orthosis in the Plan Treatment listed. |
| **Repetition** | Subsequently EEG was noted to have no seizure activity. | Subsequently EEG was noted to have no seizure seizure activity. |
| **RWS** | Patient had some discomfort but was able to tolerate procedure. | Patient had some discomfort but was able to tolerate therapy. |
| **SPV** | The patient has been given a stool softener (Senokot). | The patient have been given a stool softener (Senokot). |
| **Verb tense** | Initial evaluation revealed corticate posturing was noted by the trauma team. | Initial evaluation reveals corticate posturing was noted by the trauma team. |

NER: Named Entity Recognition, RE: Relation Extraction, TI: Textual Inference, SS: Semantic Similarity, LCC: Letter Case Changing, CMW: Common Misspelled Words, RWA: Replacement With Abbreviation, AE: Abbreviation Expansion, RWS: Replacement With Synonym, SPV: Singular/Plural Verb.

## 5. Results and Discussion

Table 3 presents the results of evaluating the neural language models on the four clinical text processing tasks. As can be seen, ClinicalXLNet generally obtained the highest results, and ClinicalELMo performed worse than the other systems. However, discussing how well the language models can perform on the benchmarks is out of the scope of this paper; we refer the reader to the respective papers for extensive discussion [10-13].

**Table 3.** Performance of the neural language models on the original test sets, character-level, and world-level perturbations to the clinical text processing tasks.

|  | ClinicalBERT | | | | ClinicalXLNet | | | | ClinicalELMo | | | |
| --- | --- | --- | --- | --- | --- | --- | --- | --- | --- | --- | --- | --- |
|  | NER | RE | TI | SS | NER | RE | TI | SS | NER | RE | TI | SS |
| **Original Test set** | | | | | | | | | | | | |
|  | 73.61 | 73.17 | 82.35 | 83.71 | 77.68 | 74.06 | 85.12 | 85.88 | 69.44 | 69.49 | 78.51 | 77.37 |
| **Character-level** | | | | | | | | | | | | |
| Deletion | 67.85 | 69.38 | 76.79 | 74.94 | 70.57 | 69.85 | 79.31 | 76.66 | 64.13 | 65.46 | 74.84 | 71.49 |
| Insertion | 68.08 | 69.86 | 76.80 | 75.95 | 72.04 | 70.57 | 79.57 | 77.92 | 64.11 | 66.54 | 74.14 | 71.05 |
| LCC | 67.23 | 67.97 | 75.17 | 76.83 | 69.31 | 68.36 | 77.05 | 77.91 | 65.85 | 66.77 | 73.02 | 72.67 |
| CMW | 66.90 | 65.83 | 73.92 | 72.60 | 68.22 | 67.63 | 76.04 | 73.92 | 63.57 | 63.23 | 72.11 | 68.95 |
| Repetition | 68.25 | 68.53 | 76.11 | 76.08 | 71.43 | 69.89 | 78.06 | 77.23 | 65.10 | 65.30 | 74.35 | 72.78 |
| Replacement | 67.44 | 69.18 | 77.05 | 76.49 | 70.07 | 69.77 | 79.48 | 77.10 | 63.81 | 64.87 | 74.40 | 71.23 |
| Swapping | 66.72 | 69.63 | 76.40 | 76.32 | 70.69 | 68.48 | 79.17 | 78.29 | 63.75 | 65.22 | 73.76 | 70.38 |
| **Word-level** | | | | | | | | | | | | |
| RWA | 67.05 | 67.58 | 75.19 | 73.85 | 70.61 | 68.19 | 77.56 | 75.42 | 61.48 | 62.64 | 71.72 | 68.70 |
| AE | 69.30 | 68.28 | 77.62 | 77.57 | 72.90 | 69.52 | 79.02 | 78.17 | 65.69 | 64.21 | 74.59 | 71.37 |
| Deletion | 68.06 | 69.03 | 78.30 | 75.80 | 72.89 | 70.02 | 79.93 | 76.63 | 63.26 | 64.51 | 74.56 | 69.14 |
| Negation | 65.91 | 64.95 | 70.03 | 71.03 | 68.39 | 65.68 | 72.31 | 73.51 | 61.43 | 59.45 | 63.29 | 65.06 |
| Word order | 60.58 | 66.34 | 75.54 | 74.11 | 63.58 | 71.97 | 80.34 | 78.11 | 57.58 | 63.24 | 67.97 | 67.51 |
| Repetition | 65.77 | 68.80 | 78.96 | 77.81 | 69.02 | 70.03 | 81.01 | 79.07 | 62.51 | 64.70 | 75.46 | 72.57 |
| RWS | 65.49 | 69.32 | 77.82 | 75.26 | 63.61 | 66.56 | 75.74 | 72.38 | 60.37 | 62.11 | 69.22 | 66.02 |
| SPV | 68.76 | 66.63 | 76.47 | 77.15 | 71.38 | 67.59 | 78.89 | 78.75 | 64.14 | 62.62 | 73.97 | 73.08 |
| Verb tense | 68.13 | 68.95 | 77.25 | 76.23 | 71.02 | 68.65 | 79.08 | 77.26 | 64.35 | 63.67 | 73.61 | 71.24 |

NER: Named Entity Recognition, RE: Relation Extraction, TI: Textual Inference, SS: Semantic Similarity, LCC: Letter Case Changing, CMW: Common Misspelled Words, RWA: Replacement With Abbreviation, AE: Abbreviation Expansion, RWS: Replacement With Synonym, SPV: Singular/Plural Verb.

Table 3 also presents the evaluation scores obtained by the neural language models on the perturbed samples from the clinical NLP datasets. In the experiments, a parameter named Perturbation Per Sample (PPS) controlled the maximum number of perturbations per sample. For brevity reasons, only the results for PPS=1 are reported in Table 3. As the results show, the performance of the language models decreases on all the clinical NLP tasks when the input contains noise. Table 4 presents the average absolute decrease in the performance of the language models on the different clinical NLP tasks when inputs contain word-level or character-level perturbations. The absolute decrease is presented for different values of PPS in the range [1, 4]. Multiple perturbations in a sample were applied by the same perturbation method. As the results demonstrate, the NLP systems are more sensitive to the word-level noise than the character-level ones. The noisy samples had more negative impact on the semantic similarity task, probably because the perturbed text changes the numerical representation of the input, consequently, the NLP system estimates different similarity score (based on the new representation) from the score it estimated for the original input. On the other hand, the relation extraction task suffers less than the other tasks from the noisy inputs, possibly because the system makes a decision based on presence of specific concepts in the text. These concepts have the most impact on the output; the system still makes a correct decision if important words that refer to important entities remain unperturbed.

**Table 4.** The average absolute decrease in the performance of the neural language models on the character-level and word-level perturbations.

| Task | AI system | Performance on test set | Character-level noise | | | | Word level noise | | | |
|---|---|---|---|---|---|---|---|---|---|---|
| | | | PPS=1 | PPS=2 | PPS=3 | PPS=4 | PPS=1 | PPS=2 | PPS=3 | PPS=4 |
| NER | ClinicalBERT | 73.61 | −6.11 | −9.15 | −10.79 | −12.03 | −7.27 | −9.20 | −10.83 | −12.04 |
| | ClinicalXLNet | 77.68 | −7.34 | −9.40 | −11.23 | −12.18 | −8.30 | −10.12 | −11.75 | −12.47 |
| | ClinicalELMo | 69.44 | −5.10 | −7.79 | −9.25 | −10.60 | −7.12 | −8.78 | −10.02 | −12.25 |
| RE | ClinicalBERT | 73.17 | −4.54 | −7.81 | −8.92 | −10.05 | −5.62 | −7.32 | −9.17 | −10.62 |
| | ClinicalXLNet | 74.06 | −4.83 | −7.03 | −9.01 | −10.36 | −5.14 | −8.19 | −10.30 | −11.43 |
| | ClinicalELMo | 69.49 | −4.14 | −6.57 | −8.03 | −9.14 | −6.47 | −8.37 | −9.96 | −11.16 |
| TI | ClinicalBERT | 82.35 | −6.31 | −8.85 | −10.20 | −12.19 | −6.21 | −8.02 | −9.75 | −11.24 |
| | ClinicalXLNet | 85.12 | −6.73 | −8.07 | −9.31 | −10.85 | −6.79 | −7.80 | −9.07 | −10.77 |
| | ClinicalELMo | 78.51 | −4.70 | −6.02 | −8.11 | −9.70 | −6.91 | −8.52 | −10.68 | −13.15 |
| SS | ClinicalBERT | 83.71 | −8.10 | −10.25 | −11.88 | −13.05 | −8.50 | −11.52 | −12.81 | −13.40 |
| | ClinicalXLNet | 85.88 | −8.87 | −9.96 | −11.59 | −12.87 | −9.06 | −11.45 | −12.70 | −13.87 |
| | ClinicalELMo | 77.37 | −6.14 | −8.22 | −9.70 | −10.61 | −7.96 | −10.96 | −13.15 | −15.33 |

NER: Named Entity Recognition, RE: Relation Extraction, TI: Textual Inference, SS: Semantic Similarity, PPS: Perturbation Per Sample.

In order to investigate whether the performance scores significantly change between different fine-tuning runs, we performed three separate fine-tuning runs (in addition to the first run reported in Table 3 and Table 4) on the same training sets for every NLP model. We then experimented the fine-tuned models with every perturbation method. Using a *paired t-test*, we assessed the variability of scores between the first run and the additional three runs. Considering a confidence interval of 95%, the *two-tailed P-value* was equal to 0.5438, 0.5416, and 0.5502 when the first run was compared against the second, third, and forth run, respectively. According to the *P-values*, the difference between the first fine-tuning run and the other three runs was not statistically significant.

It can be seen from the results that some NLP systems can handle specific types of noise more effectively than the others. For instance, ClinicalELMo is less sensitive to character-level noise. This may be an effect of its pure character-based representations helping the model to use morphological clues when dealing with character-level noise. ClinicalXLNet is shown to be more robust to **Word ordering** noise. This may have resulted from the permutation language modelling strategy that enables the NLP system to capture the context by observing important words, regardless of their position in the text. The results also reveal that ClinicalBERT can deal with the **Replace with synonym** noise more efficiently than the other systems, probably because it was pretrained on larger corpora.

Figure 1 shows six noisy samples from the clinical NLP datasets. ClinicalBERT produced correct outputs for the original samples, but made wrong decisions for the noisy inputs. In Example 1, one character removed from both words 'Electroencephalogram' and 'seizure'. The text still seems understandable and the NLP system is expected to find the right relation between the medical problem and treatment mentioned in the text, but the minor character-level noise caused the system to produce a wrong output. Example 2 contains two misspelled words that still seem understandable, but ClinicalBERT could not estimate the right similarity score for the pair of sentences. In Example 3, minor character-level swapping noise changed the word 'symptoms' to 'symtpoms' and 'severe' to 'seveer'. This noise does not change the semantics, but the NLP system wrongly decided no contradiction relationship exists between the two sentences.

In Example 4, two phrases have been replaced with their abbreviations; shortness of breath has been replaced with SOB, and chest pain has been replaced with CP. The semantics do not change but ClinicalBERT could not estimate the correct similarity score it estimated for the original, unperturbed sample. In Example 5, two words, i.e. 'initial' and 'trauma', were removed but the relation between the test and the medical problem is still recognizable. However, the NLP system was not able to produce the correct outcome. In Example 6, the second sentence is negated by converting the positive verb 'showed' to it negative form 'did not show'. The meaning has changed and the NLP system is expected to find the contradiction, but it could not adapt its output to reflect the injected negation.

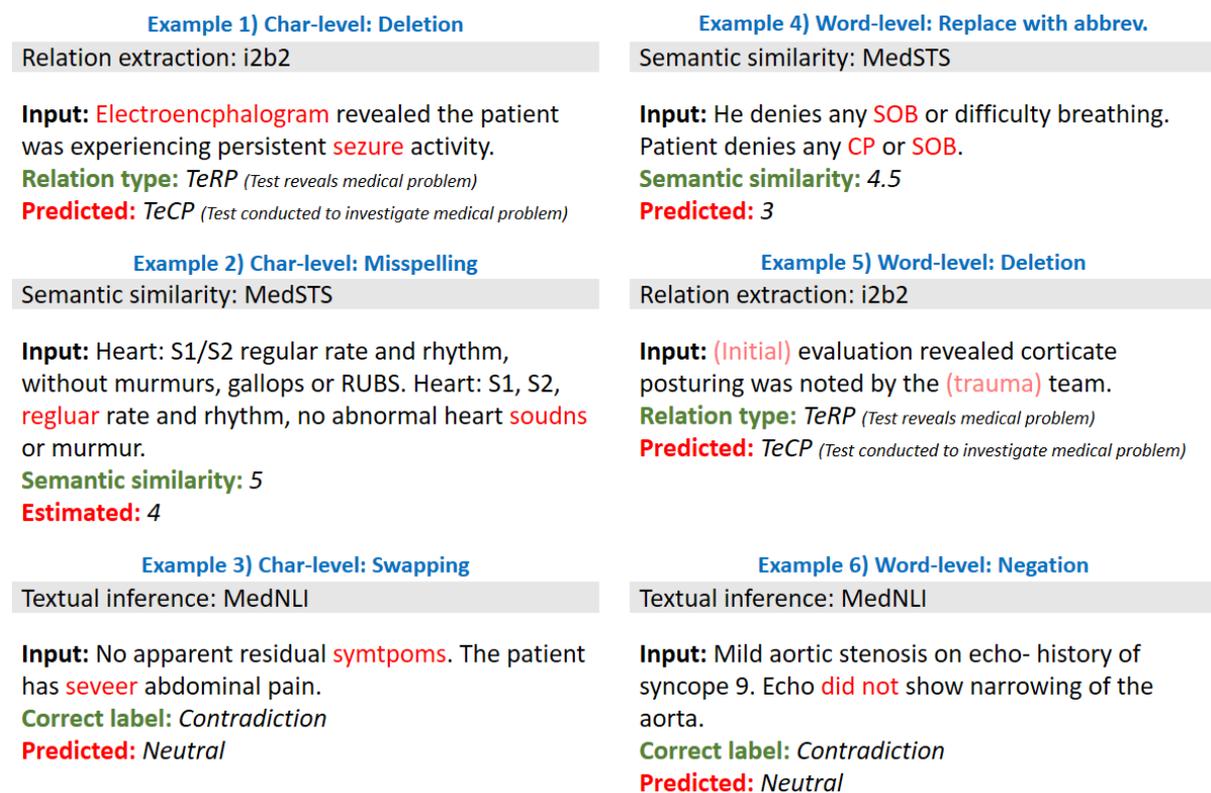

**Figure 1.** Noisy samples from the clinical NLP datasets. ClinicalBERT produced correct outputs for the original samples, but made wrong decisions for the noisy inputs.

## 6. User study

We conducted a user study with 10 participants to investigate whether perturbed samples can be automatically produced and used to evaluate the robustness of clinical NLP systems, or if persistent human supervision is needed to make sure noisy samples are still meaningful and understandable. Every participant either had completed a medical degree or was studying in a medical program at a university and had sufficient medical knowledge to judge the understandability and meaningfulness of perturbed samples. Every participant was given a questionnaire containing 50 noisy samples from the datasets used in our experiments. The questionnaire covered all the character-level and word-level perturbation methods introduced in this paper. Using the *Fleiss' kappa measure*, we assessed the inter-rater agreement among the participants. The calculated *kappa value* was equal to 0.5054 that can be interpreted as moderate agreement among the participants, according to Landis and Koch [33].

In the first part of the questionnaire, the users were given 30 noisy samples from the perturbation methods that are expected to change text meaning only infrequently. These perturbation methods were all the character-level ones and five word-level perturbations, i.e. **Replacement with abbreviation**, **Abbreviation expansion**, **Repetition**, **Singular/plural verb**, and **Verb tense**. For every perturbed sample, the participants were asked to judge if the noisy text was still understandable and conveys the same meaning as the original text. According to a majority voting evaluation, the participants reported that 87% of the perturbed samples were understandable and had the same meaning as the original text. For 10% of the samples, there was a tie and it was not possible to decide whether the perturbed text preserved the meaning or not through majority voting. The remaining 3% were judged to be understandable but having a changed meaning. These results demonstrate sufficient preservation of text meaning to use these methods in automated pipelines for producing noisy samples.

In the second part of the questionnaire, the users were given 20 noisy samples from perturbation methods that were expected to change text meaning more frequently. These were the remaining word-level perturbations, i.e. **Deletion**, **Negation**, **Word order**, and **Replacement with synonym**. For every perturbed sample, the participants were asked to judge if the noisy text was still understandable and conveys the same meaning as the original text. According to a majority voting evaluation, the participants reported that 30% of the perturbed samples were understandable and had the same meaning as the original text, 25% were still understandable but the meaning changed, 40% were not understandable anymore, and there was a tie for 5% of the samples. These results demonstrate that noisy samples produced by these perturbation methods should be monitored, filtered, or corrected to ensure the noisy text is understandable, meaningful, and consistent with the ground-truth label.

It is worth to note that deciding which perturbation methods may change the text's meaning depends on the task at hand. For example, changing verb tense may change the input's meaning in a question answering task and the NLP system is expected to produce a different answer, whereas, this perturbation is not expected to affect the NLP system's decision in a NER task where disease names are identified. Therefore, we decided which perturbation methods may change the meaning with respect to the NLP tasks we used in the experiments. This choice may differ when other tasks are investigated.

## 7. Conclusion

In this paper, we introduced and implemented various types of character-level and word-level perturbation methods that can be utilized to simulate real-world situations in which clinical text processing systems encounter noisy inputs. We also evaluated the robustness of different high-performance NLP systems on several clinical text processing tasks. The results demonstrated that state-of-the-art clinical NLP models make erroneous decisions when the input is slightly noisy. This stimulates future research on improving the robustness and reliability of AI systems that are developed and tested for being used as clinical decision support tools.

The user studies showed many perturbation methods can be used automatically to generate noisy test samples, but some produce samples that need to be manually validated to ensure meaningfulness and consistency. This allows system developers, software testers, and end-users of clinical NLP tools to design and implement automatic and semi-automatic tests that reveal vulnerabilities of their systems.

## References

[1] Y. LeCun, Y. Bengio, and G. Hinton, "Deep learning," *Nature,* vol. 521, pp. 436-444, 2015.
[2] S. Min, B. Lee, and S. Yoon, "Deep learning in bioinformatics," *Briefings in Bioinformatics,* vol. 18, pp. 851-869, 2017.
[3] K. Lan, D.-t. Wang, S. Fong, L.-s. Liu, K. K. L. Wong, and N. Dey, "A Survey of Data Mining and Deep Learning in Bioinformatics," *Journal of Medical Systems,* vol. 42, p. 139, 2018.
[4] Y. Yu, M. Li, L. Liu, Y. Li, and J. Wang, "Clinical big data and deep learning: Applications, challenges, and future outlooks," *Big Data Mining and Analytics,* vol. 2, pp. 288-305, 2019.
[5] D. Ravì, C. Wong, F. Deligianni, M. Berthelot, J. Andreu-Perez, B. Lo*, et al.*, "Deep Learning for Health Informatics," *IEEE Journal of Biomedical and Health Informatics,* vol. 21, pp. 4-21, 2017.
[6] S. Srivastava, S. Soman, A. Rai, and P. K. Srivastava, "Deep learning for health informatics: Recent trends and future directions," in *2017 International Conference on Advances in Computing, Communications and Informatics (ICACCI)*, 2017, pp. 1665-1670.
[7] B. Shickel, P. J. Tighe, A. Bihorac, and P. Rashidi, "Deep EHR: A Survey of Recent Advances in Deep Learning Techniques for Electronic Health Record (EHR) Analysis," *IEEE Journal of Biomedical and Health Informatics,* vol. 22, pp. 1589-1604, 2018.
[8] Y. Wang, L. Wang, M. Rastegar-Mojarad, S. Moon, F. Shen, N. Afzal*, et al.*, "Clinical information extraction applications: A literature review," *Journal of Biomedical Informatics,* vol. 77, pp. 34-49, 2018.
[9] S. Wu, K. Roberts, S. Datta, J. Du, Z. Ji, Y. Si*, et al.*, "Deep learning in clinical natural language processing: a methodical review," *Journal of the American Medical Informatics Association,* vol. 27, pp. 457-470, 2019.

[10] Y. Peng, S. Yan, and Z. Lu, "Transfer Learning in Biomedical Natural Language Processing: An Evaluation of BERT and ELMo on Ten Benchmarking Datasets," in *Proceedings of the BioNLP 2019 workshop*, Florence, Italy, 2019, pp. 58-65.

[11] Y. Gu, R. Tinn, H. Cheng, M. Lucas, N. Usuyama, X. Liu*, et al.*, "Domain-specific language model pretraining for biomedical natural language processing," *arXiv preprint arXiv:2007.15779,* 2020.

[12] E. Alsentzer, J. Murphy, W. Boag, W.-H. Weng, D. Jindi, T. Naumann*, et al.*, "Publicly Available Clinical BERT Embeddings," in *Proceedings of the 2nd Clinical Natural Language Processing Workshop*, Minneapolis, Minnesota, USA, 2019, pp. 72-78.

[13] K. Huang, A. Singh, S. Chen, E. Moseley, C.-Y. Deng, N. George*, et al.*, "Clinical XLNet: Modeling Sequential Clinical Notes and Predicting Prolonged Mechanical Ventilation," in *Proceedings of the 3rd Clinical Natural Language Processing Workshop*, Online, 2020, pp. 94-100.

[14] M. T. Ribeiro, T. Wu, C. Guestrin, and S. Singh, "Beyond Accuracy: Behavioral Testing of NLP Models with CheckList," in *Proceedings of the 58th Annual Meeting of the Association for Computational Linguistics*, Online, 2020, pp. 4902-4912.

[15] K. Ethayarajh and D. Jurafsky, "Utility is in the Eye of the User: A Critique of NLP Leaderboards," in *Proceedings of the 2020 Conference on Empirical Methods in Natural Language Processing (EMNLP)*, Online, 2020, pp. 4846-4853.

[16] R. Leaman, R. Khare, and Z. Lu, "Challenges in clinical natural language processing for automated disorder normalization," *Journal of Biomedical Informatics,* vol. 57, pp. 28-37, 2015.

[17] X. Yang, X. He, H. Zhang, Y. Ma, J. Bian, and Y. Wu, "Measurement of Semantic Textual Similarity in Clinical Texts: Comparison of Transformer-Based Models," *JMIR Med Inform,* vol. 8, p. e19735, 2020.

[18] A. Holzinger, J. Schantl, M. Schroettner, C. Seifert, and K. Verspoor, "Biomedical Text Mining: State-of-the-Art, Open Problems and Future Challenges," in *Interactive Knowledge Discovery and Data Mining in Biomedical Informatics: State-of-the-Art and Future Challenges*, A. Holzinger and I. Jurisica, Eds., ed Berlin, Heidelberg: Springer Berlin Heidelberg, 2014, pp. 271-300.

[19] S. Sahu, A. Anand, K. Oruganty, and M. Gattu, "Relation extraction from clinical texts using domain invariant convolutional neural network," in *Proceedings of the 15th Workshop on Biomedical Natural Language Processing*, Berlin, Germany, 2016, pp. 206-215.

[20] G. Zeng, F. Qi, Q. Zhou, T. Zhang, B. Hou, Y. Zang*, et al.*, "OpenAttack: An Open-source Textual Adversarial Attack Toolkit," *arXiv preprint arXiv:2009.09191,* 2020.

[21] W. E. Zhang, Q. Z. Sheng, A. Alhazmi, and C. Li, "Adversarial Attacks on Deep-learning Models in Natural Language Processing: A Survey," *ACM Trans. Intell. Syst. Technol.,* vol. 11, p. Article 24, 2020.

[22] S. J. Pan and Q. Yang, "A Survey on Transfer Learning," *IEEE Transactions on Knowledge and Data Engineering,* vol. 22, pp. 1345-1359, 2010.

[23] M. Peters, M. Neumann, M. Iyyer, M. Gardner, C. Clark, K. Lee*, et al.*, "Deep Contextualized Word Representations," in *Proceedings of the 2018 Conference of the North American Chapter of the Association for Computational Linguistics: Human Language Technologies*, New Orleans, Louisiana, 2018, pp. 2227-2237.

[24] B. Nye, J. Jessy Li, R. Patel, Y. Yang, I. J. Marshall, A. Nenkova*, et al.*, "A Corpus with Multi-Level Annotations of Patients, Interventions and Outcomes to Support Language Processing for Medical Literature," *Proceedings of the conference. Association for Computational Linguistics. Meeting,* vol. 2018, pp. 197-207, 2018.

[25] Ö. Uzuner, B. R. South, S. Shen, and S. L. DuVall, "2010 i2b2/VA challenge on concepts, assertions, and relations in clinical text," *Journal of the American Medical Informatics Association,* vol. 18, pp. 552-556, 2011.

[26]     A. Romanov and C. Shivade, "Lessons from Natural Language Inference in the Clinical Domain," in *Proceedings of the 2018 Conference on Empirical Methods in Natural Language Processing*, Brussels, Belgium, 2018, pp. 1586-1596.
[27]     Y. Wang, S. Fu, F. Shen, S. Henry, O. Uzuner, and H. Liu, "The 2019 n2c2/OHNLP Track on Clinical Semantic Textual Similarity: Overview," *JMIR Med Inform,* vol. 8, p. e23375, 2020.
[28]     (accessed 01/03/2021). *List of medical abbreviations*. <https://en.wikipedia.org/wiki/List_of_medical_abbreviations>
[29]     (accessed 01/03/2021). *67 Medical Abbreviations and Acronyms Every Caregiver Should Know*. <https://seniorsresourceguide.com/articles/art01112.html>
[30]     (accessed 01/03/2021). *Medical Abbreviations List*. <https://abbreviations.yourdictionary.com/articles/medical-abbrev.html>
[31]     (accessed 01/03/2021). *Common Medical Abbreviations List*. <https://www.medicinenet.com/common_medical_abbreviations_and_terms/article.htm>
[32]     (accessed 01/12/2020). *The SketchEngine Language Tool*. <https://www.sketchengine.eu/>
[33]     J. R. Landis and G. G. Koch, "The Measurement of Observer Agreement for Categorical Data," *Biometrics,* vol. 33, pp. 159-174, 1977.